# Text Sentiment Analysis and Classification Based on Bidirectional Gated Recurrent Units (GRUs) Model

**Wei Xu[1*], Jianlong Chen[1], Zhicheng Ding[2], Jinyin Wang[3]**

[1]Independent researcher, Los Altos, 94024, USA
[1]Independent researcher, Beijing, 100029, China
[2]Independent researcher, Renton, 98059, USA
[3]Independent researcher, Jersey City, 07306, USA
*Corresponding author email:williamxw09@gmail.com

**Abstract.** This paper explores the importance of text sentiment analysis and classification in the field of natural language processing, and proposes a new approach to sentiment analysis and classification based on the bidirectional gated recurrent units (GRUs) model. The study firstly analyses the word cloud model of the text with six sentiment labels, and then carries out data preprocessing, including the steps of removing special symbols, punctuation marks, numbers, stop words and non-alphabetic parts. Subsequently, the data set is divided into training set and test set, and through model training and testing, it is found that the accuracy of the validation set is increased from 85% to 93% with training, which is an increase of 8%; at the same time, the loss value of the validation set decreases from 0.7 to 0.1 and tends to be stable, and the model is gradually close to the actual value, which can effectively classify the text emotions. The confusion matrix shows that the accuracy of the model on the test set reaches 94.8%, the precision is 95.9%, the recall is 99.1%, and the F1 score is 97.4%, which proves that the model has good generalisation ability and classification effect. Overall, the study demonstrated an effective method for text sentiment analysis and classification with satisfactory results.

**Keywords:** Text Sentiment Analysis, Bidirectional gated recurrent units, Classification.

## 1. Introduction

Text Sentiment Analysis and Classification is an important research direction in the field of Natural Language Processing, aiming at automatically identifying and understanding the emotional colours embedded in texts through computer technology, and then classifying the sentiments of texts. The research background of this field can be traced back to the in-depth exploration of human language understanding and emotion expression [1]. With the generation of large amounts of text data such as social media, online comments, news reports, etc., it is hoped that computer technology can be used to more quickly and accurately analyse and mine the emotional tendencies and attitudes embedded in this information to help in decision-making, public opinion monitoring, and marketing.

In our upcoming research endeavors, we aim to integrate Named Entity Recognition (NER) to optimize text classification. Furthermore, we will delve into the effectiveness of both BERT and CRF models, as demonstrated in the study conducted by Liu et al., and conduct a comparative analysis of their performance against that of the GRU model [2,3]. And deep learning, as a machine learning method based on neural network models, plays an important role in text sentiment analysis and classification. Deep learning algorithms can automatically learn feature representations from data with higher levels of abstraction and expressiveness, avoiding the limitations imposed by manually designed features.

GRU has numerous applications in natural language processing (NLP) [4] and Recurrent Neural Network (RNN) [5] are two mainstream models commonly used in text sentiment analysis and classification tasks.CNN can capture local features when processing text and extract different levels of semantic information through multi-layer convolutional operations, while RNN is suitable for processing sequential data, and is better able to capture contextual information. In addition, structures such as Long Short-Term Memory Networks (LSTM) and Gated Recurrent Units (GRU) are widely used to solve problems such as gradient vanishing or gradient explosion during RNN training.

In addition to the models mentioned above, the attention mechanism is also one of the key techniques that have been widely used in text sentiment analysis and classification tasks in recent years. By introducing the attention mechanism, the model can be more flexible to focus on the information at different locations in the input sequence, thus improving the model's effectiveness in processing long texts [6].

Deep learning algorithms have better results in the task of text sentiment analysis and classification, and this paper adds a new idea for text sentiment analysis and classification based on the bidirectional gated recurrent units (GRUs) model. With the continuous development and improvement of technology, it is believed that more innovative methods will emerge in the future, bringing more possibilities and development space for this field.

## 2. Data sources and statistics

The data used in this paper is a text labelling dataset, labelled and produced by several experts, totalling 461810 text messages and their corresponding label classifications. The dataset contains Text and its corresponding emotions, the six emotions are anger, fear, joy, love, sadness and surprise. Part of the dataset is shown in Table 1. Where 0 denotes Sadness, 1 denotes Joy, 2 denotes Love, 3 denotes Anger, 4 denotes Fear and 5 denotes Surprise and each text has its corresponding emotion label.

**Table 1.** Partial data.

| Text | Label |
|---|---|
| I just feel really helpless and heavy hearted | 4 |
| I ve enjoyed being able to slouch about relax and unwind and frankly needed it after those last few weeks around the end of uni and the expo i have lately started to find myself feeling a bit listless which is never really a good thing | 0 |
| I gave up my internship with the dmrg and am feeling distraught | 4 |
| I dont know i feel so lost | 0 |
| I am a kindergarten teacher and i am thoroughly weary of my job after having taken the university entrance exam i suffered from anxiety for weeks as i did not want to carry on with my work studies were the only alternative | 4 |

## 3. Word cloud model

The word cloud model presents high-frequency words in a visual way by statistically analysing the frequency of words in the text. It first needs to process the text with word segmentation to split the text into individual words. Then, the frequency of each word in the text is counted, and the size and colour shade of the words in generating the word cloud map are determined according to the frequency magnitude.

The texts with six emotion labels were analysed separately to produce the word cloud model, and the results are shown in Figure 1.

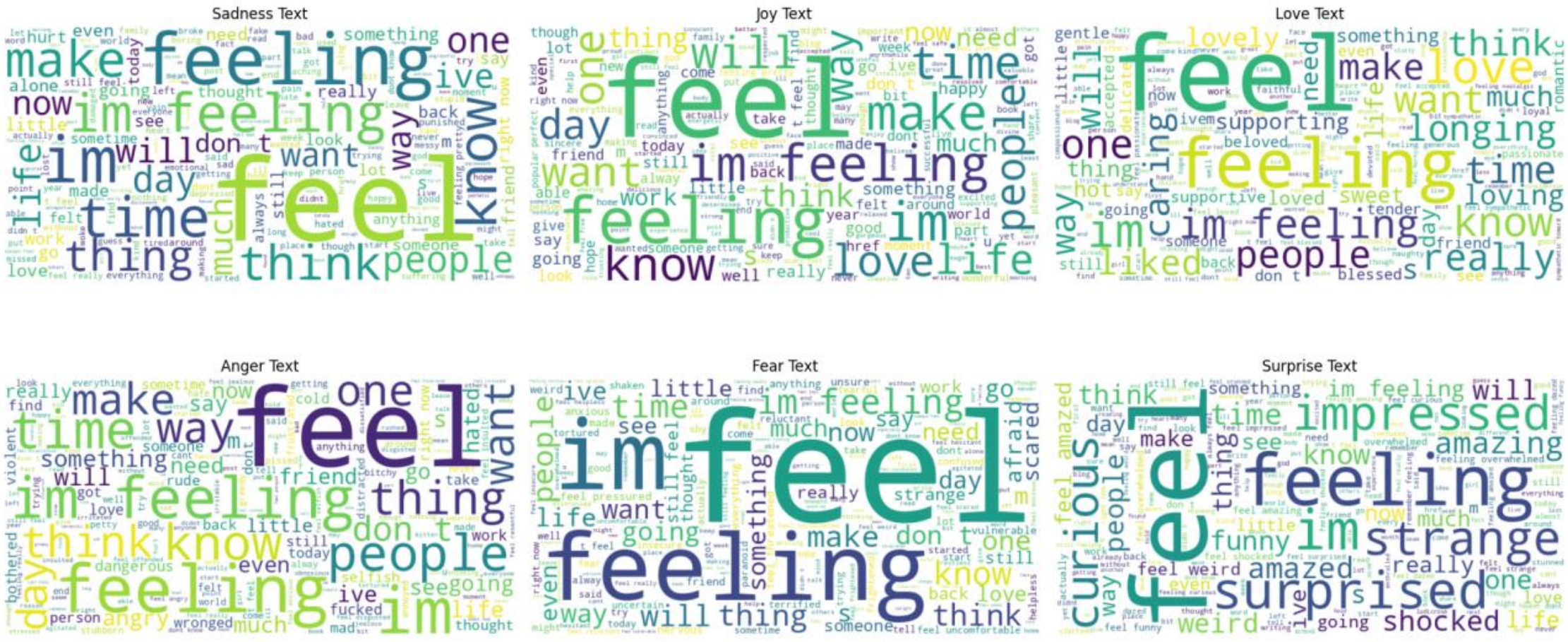

**Figure 1.** Word cloud model.
（Photo credit : Original）

## 4. Text processing

Text data preprocessing is a very important step in natural language processing, which can help to improve the quality and accuracy of text data. In this paper, the preprocessing of text data includes the steps of deleting special symbols and punctuation marks, deleting numerical values in text, deleting stop words and deleting non-alphabetic parts.

(1) Special symbols and punctuation usually do not contain useful semantic information, and deleting them can simplify text data and reduce noise interference. Removing special symbols and punctuation marks helps to reduce the complexity of text data, making subsequent text analysis more efficient and accurate.

(2) Numerical values usually do not affect the semantic content of text data, and removing them can avoid interfering with natural language processing tasks. Removing numerical values can make the text more pure and facilitate subsequent text analysis tasks such as word frequency statistics and topic modelling.

(3) Discontinued words are words that appear frequently in natural language but usually do not help much in understanding the whole text, such as "the", "is", etc. Removing them can reduce the number of words. Removing deactivated words can reduce the size of the bag-of-words model and improve the differentiation of the text feature vocabulary, thus enhancing the effectiveness of the natural language processing task.

(4) Non-alphabetic parts include numbers, special symbols, etc., which may interfere with the natural language processing task in some cases. Removing the non-alphabetic parts helps to retain the parts of the text that really have semantic information, making the subsequent processing more focused on the linguistic content rather than other irrelevant factors.

Through the above preprocessing steps, the original text data can be effectively simplified and optimised, laying a good foundation for the next text mining, classification, clustering and other tasks.

## 5. Bi-directional gated cycle unit model

The bidirectional gated recurrent unit model is a model commonly used in the field of neural networks for processing sequence data and time series data [7]. The model combines the features of Recurrent Neural Networks (RNN) and gating mechanism, which can effectively capture long-term dependencies and has the advantages of strong memory and good adaptability. The structure of the bidirectional gated recurrent unit model is shown in Figure 2.

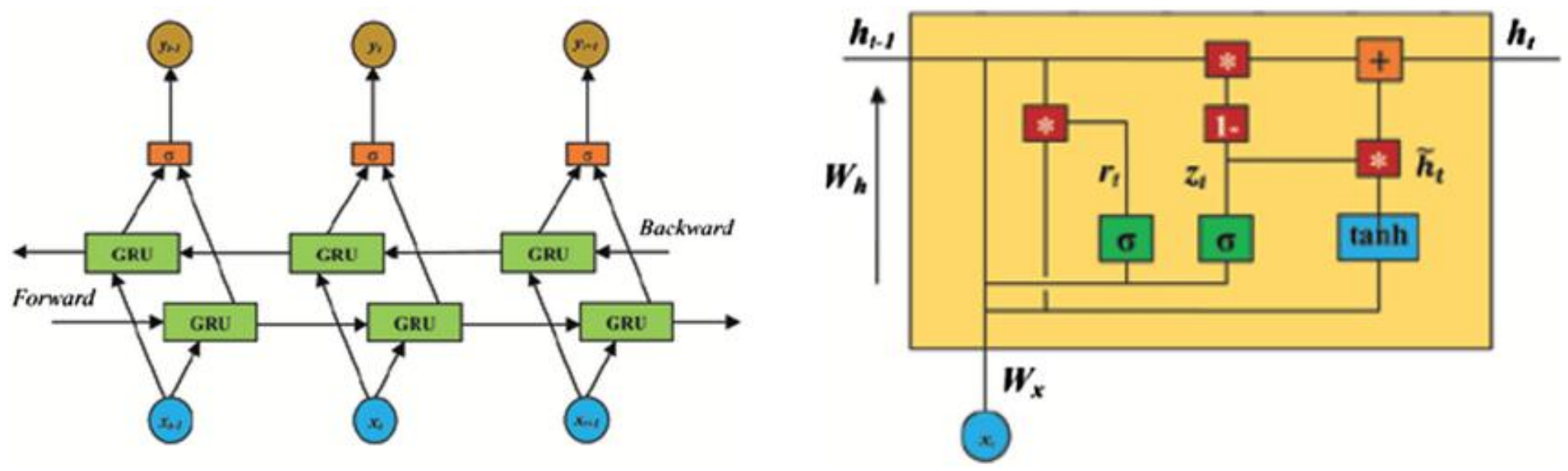

**Figure 2.** The structure of the bidirectional gated recurrent unit model.
（Photo credit : Original）

Firstly, the bidirectional gated cyclic unit model contains cyclic units in two directions: one is a forward cyclic unit and the other is a reverse cyclic unit [8]. Each of these two directions is responsible for capturing the dependencies between the preceding and following information in the input sequence, leading to a more comprehensive understanding of the entire sequence. At each time step, both the forward loop unit and the reverse loop unit receive the input of the current moment as well as the hidden state of the previous moment, and output the hidden state of the current moment.

Second, the two-way gated loop cell model introduces gating mechanisms, which mainly include forgetting gates, input gates and output gates. These gating units control the flow of information through a sigmoid function, which can effectively filter irrelevant information and retain important information [9]. The forgetting gate is used to determine how much information in the hidden state at the previous moment needs to be forgotten; the input gate is used to determine how much information in the new input data at the current moment needs to be added to the hidden state; and the output gate is used to regulate which information in the hidden state needs to be exported to the next layer or output layer.

The bi-directional gated cyclic cell model is also characterised by strong memory [10]. By constantly updating the hidden states and passing information, the model can better preserve historical information and extract the relevant content when needed for prediction or classification tasks. Meanwhile, due to the introduction of the gating mechanism, the problem of gradient vanishing or gradient explosion is less likely to occur when dealing with long sequence data, which improves the stability and training effect of the model.

## 6. Results

The summary list of models is shown in Table 2, and the experiments were run using python 3.10 with a training and test set ratio of 8:2, and a total of 5 epochs were trained.

**Table 2.** Partial data.

| Layer(type) | Output Shape | Param# |
|---|---|---|
| Embedding 2(Embedding) | (None, 79, 50) | 2500000 |
| Dropout 1(Dropout) | (None, 79, 50) | 0 |
| Bidirectional 6(Bidirectional) | (None, 79, 240) | 123840 |
| Bidirectional 7(Bidirectional) | (None, 79, 128) | 117504 |
| Batch normalization 2(Batch Normalization) | (None, 79, 128) | 512 |
| Bidirectional 8(Bidirectional) | (None, 128) | 74496 |
| Dense 2(Dense) | (None,6) | 774 |

The changes in the values of LOSS and ACCURACY of the training and test sets are recorded and plotted as line graphs respectively, the ACCURACY of the training and test sets are shown in Fig. 3.(a) and the LOSS of the training and test sets are shown in Fig. 3.(b).

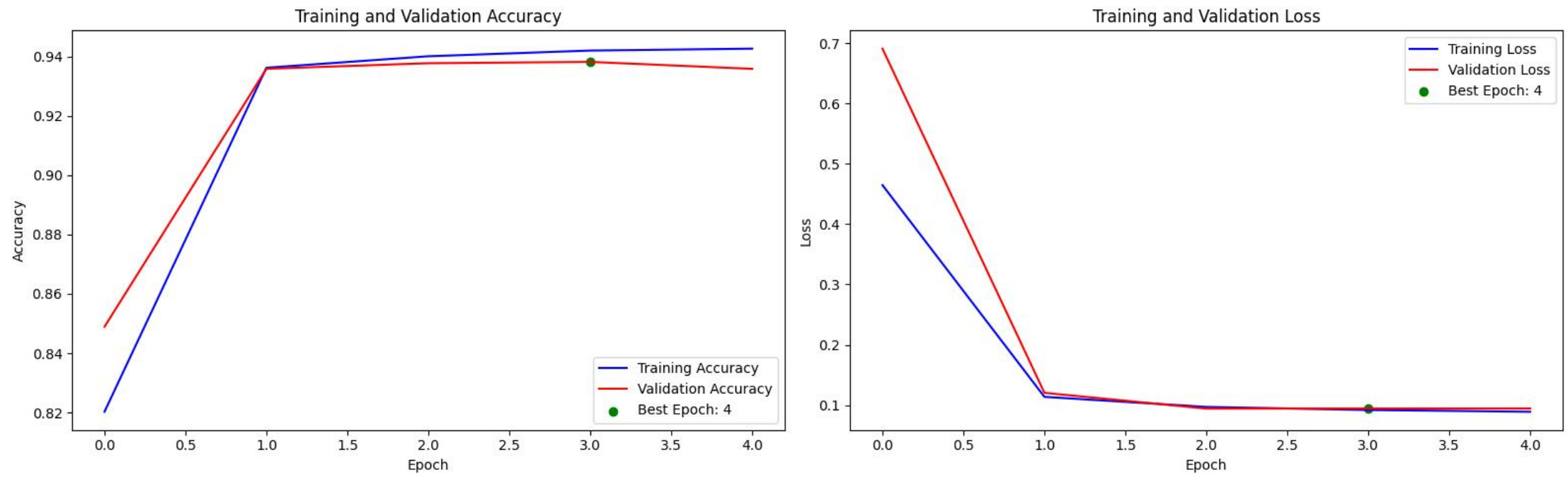

(a) (b)

**Figure 3.** The structure of the bidirectional gated recurrent unit model. (a) ACCURACY. (b) LOSS.
（Photo credit : Original）

As can be seen from the changes in the accuracy of the training and test sets, the accuracy of the validation set changes from the initial 85% to 93% as the training process proceeds, and the accuracy improves by 8%.

As can be seen from the changes in the loss of the training and test sets, as the training process proceeds, the loss of the validation set changes from 0.7 to 0.1 and tends to converge, and the predicted value of the model gradually converges to the actual value, which is able to classify the emotions of the text very well.

The model is tested using the test set, the accuracy of the test set is calculated, and the confusion matrix predicted by the test set is output, and the results are shown in Fig. 4.

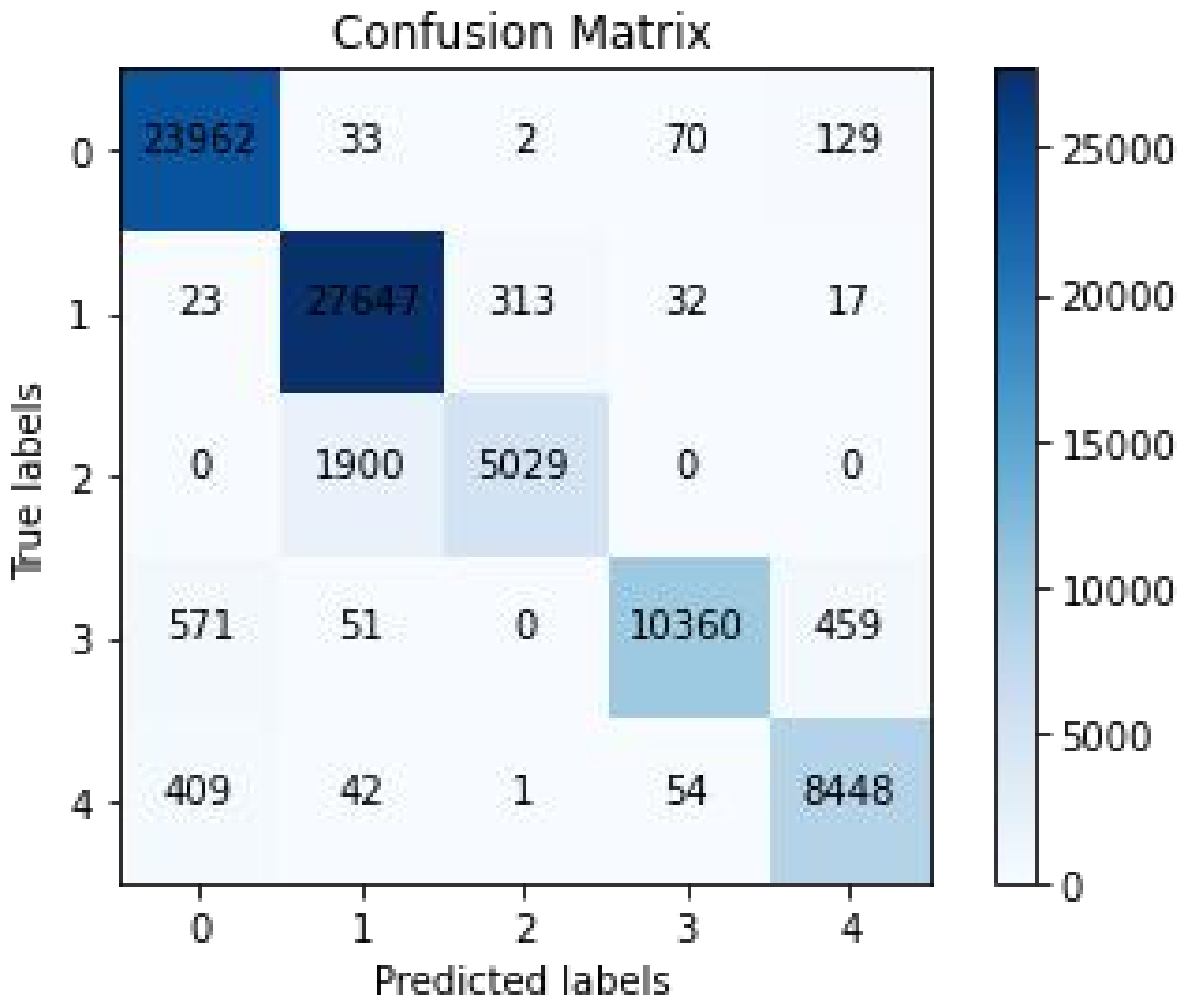

**Figure 4.** Confusion matrix.
（Photo credit : Original）

From the confusion matrix, the prediction accuracy of the model's test set is 94.8%, the precision is 95.9%, the recall is 99.1%, and the F1 score is 97.4%, the model also performs well in the test set, showing the model's excellent generalisation ability to classify textual emotions well.

## 7. Conclusion

In this paper, we conduct a research on text sentiment analysis and classification based on the bidirectional gated recurrent units (GRUs) model, which brings new ideas and methods to the field of sentiment recognition. Firstly, the text with six emotion labels is analysed by producing a word cloud model, which provides a data base for subsequent preprocessing and model training. Second, in the data preprocessing stage, a series of steps were taken, including the deletion of special symbols, punctuation marks, numerical values, stop words, and non-alphabetic parts, to ensure the purity and usability of the text data. Then, in the dataset partitioning and model training testing phase, by monitoring the trends of accuracy and loss, the accuracy of the validation set increased from 85% to 93%, which is an 8% increase in accuracy; at the same time, the loss of the validation set decreased from 0.7 to 0.1 and converged, and the model prediction was gradually close to the actual value, and the accurate classification of textual sentiment was successfully achieved.

In addition, in the confusion matrix analysis, it is found that the model performs well on the test set: the prediction accuracy reaches 94.8%, the precision reaches 95.9%, the recall reaches 99.1%, and the F1 score is as high as 97.4%. This shows that the model has excellent generalisation ability and classification effect, and is able to make accurate judgments even in the face of unknown text sentiment. Overall, the text sentiment analysis method based on the GRUs model proposed in this study shows good results and potential in the experiments, and brings new insights and application prospects to the field of sentiment recognition technology.

**References**

[1] Peng, Qucheng, et al. "RAIN: regularization on input and network for black-box domain adaptation." Proceedings of the Thirty-Second International Joint Conference on Artificial Intelligence, 2023.

[2] Liu, Ziyu, et al. "Named Entity Recognition and Named Entity on Esports Contents." 2020 15th Conference on Computer Science and Information Systems (FedCSIS), IEEE, 2020.

[3] Chen, Tin-Chih Toly, Hsin-Chieh Wu, and Min-Chi Chiu. "A deep neural network with modified random forest incremental interpretation approach for diagnosing diabetes in smart healthcare." Applied Soft Computing 152 (2024): 111183.

[4] Zhao, Jinman, et al. "Structural Realization with GGNNs." Proceedings of the Fifteenth Workshop on Graph-Based Methods for Natural Language Processing (TextGraphs-15), edited by Alexander Panchenko et al., Association for Computational Linguistics, 2021, pp. 115 – 124. https://doi.org/10.18653/v1/2021.textgraphs-1.11.

[5] Al-Tameemi, Israa K. Salman, et al. "Interpretable multimodal sentiment classification using deep multi-view attentive network of image and text data." IEEE Access (2023).

[6] Haque, Rezaul, et al. "Multi-class sentiment classification on Bengali social media comments using machine learning." International journal of cognitive computing in engineering 4 (2023): 21-35.

[7] Zou, Haochen, and Zitao Wang. "A semi-supervised short text sentiment classification method based on improved Bert model from unlabelled data." Journal of Big Data 10.1 (2023): 35.

[8] Zhang, **angsen, et al. "Text sentiment classification based on BERT embedding and sliced multi-head self-attention Bi-GRU." Sensors 23.3 (2023): 1481.

[9] Qorich, Mohammed, and Rajae El Ouazzani. "Text sentiment classification of Amazon reviews using word embeddings and convolutional neural networks." The Journal of Supercomputing 79.10 (2023): 11029-11054.

[10] Khan, Jawad, et al. "Sentiment and context-aware hybrid DNN with attention for text sentiment classification." IEEE Access 11 (2023): 28162-28179.